\pdfoutput=1

\documentclass[11pt]{article}

\usepackage{emnlp2021}

\usepackage{times}
\usepackage{latexsym}
\usepackage{bm}
\usepackage{hyperref}
\usepackage{color,colortbl}
\usepackage{stfloats}
\usepackage{subfigure}
\usepackage{enumitem}
\usepackage{booktabs}
\usepackage{nccmath}
\usepackage{multirow}
\usepackage{subfigure}
\usepackage{latexsym}
\usepackage{todonotes}
\usepackage{makecell}
\usepackage{dashrule}

\usepackage{graphicx}

\usepackage[T1]{fontenc}

\usepackage[utf8]{inputenc}

\usepackage{microtype}

\definecolor{LightGray}{gray}{0.9}
\usepackage{soul}

\title{Generic resources are what you need: \\ Style transfer tasks without task-specific parallel training data}

\author{
Huiyuan Lai, Antonio Toral, Malvina Nissim\\
CLCG, University of Groningen / The Netherlands\\
\texttt{\{h.lai, a.toral.ruiz, m.nissim\}@rug.nl}
}

\begin{document}
\maketitle

\begin{abstract}
Style transfer aims to rewrite a source text in a different target style while preserving its content. 
We propose a novel approach to this task that leverages generic resources, and without using any task-specific parallel (source--target) data
%
 outperforms existing unsupervised approaches on the two most popular style transfer tasks: formality transfer and polarity swap. In practice, we adopt a multi-step procedure which builds on a generic pre-trained sequence-to-sequence model (BART). First, we strengthen the model's ability to \textit{rewrite} by further pre-training BART on both an existing collection of generic paraphrases, as well as on synthetic pairs created using a general-purpose lexical resource. Second, through an iterative back-translation approach, we train two models, each in a transfer direction, so that they can provide each other with synthetically generated pairs, dynamically in the training process. Lastly, we let our best resulting model generate static synthetic pairs to be used in a supervised training regime.
Besides methodology and state-of-the-art results, a core contribution of this work is a reflection on the nature of the two tasks we address, and how their differences are highlighted by their response to our approach. 
\end{abstract}

\section{Introduction}
\label{sec:intro}

Text style transfer is, broadly put, the task converting a text of one style into another while preserving its content. 
In its recent tradition within Natural Language Generation (NLG), two tasks and their corresponding datasets have been commonly used \citep{Zhirui-2018, fuli-2019, wu-etal-2019-hierarchical-reinforced, xiaoyuan-ijcai, zhou-etal-2020-exploring}. One dataset was specifically created for \textit{formality transfer} and contains parallel data (GYAFC \citep{rao-tetreault-2018}), while the other one contains a large amount of non-parallel sentiment labelled texts (YELP \citep{li-etal-2018-delete}), with parallel pairs for 
test, and is used for the task of \textit{polarity swap}. Examples from these datasets are shown in Table~\ref{tab:example}.

The two tasks are usually conflated in the literature 
under the general \textit{style transfer}
label and addressed with the same methods, but we find this an oversimplification.
Formality transfer implies rewriting a formal sentence into its informal counterpart (or viceversa) while preserving its meaning. Polarity swap, instead, aims to change a positive text into a negative one (or viceversa); and while the general theme must be preserved, the meaning is by definition not maintained (e.g. ``I hated that film'' $\rightarrow$ ``I loved that film''). 
In line with previous work, 
we also address both tasks in a similar way, but this is actually to 
unveil how their different nature affects modelling and evaluation.

Due to the general scarcity of parallel data, previous works mainly adopted unsupervised approaches, dubbed \textit{unpaired}
methods \citep{dai-etal-2019-style} since they do not rely on labelled training pairs. However, it has also been shown that best results, unsurprisingly, can be achieved if parallel training data (such as the formality dataset \citep{rao-tetreault-2018}) is available \cite{Abhilasha-2020, lai-etal-2021}. For this reason, substantial work has gone into the creation of artificial training pairs through various methods (see Section~\ref{sec:related}); approaches using synthetic pairs are thus still considered unsupervised in the style transfer literature, since they do not use manually labelled data. 

We explore how parallel data can best be derived and integrated in a general style transfer framework. To do so, we create pairs in a variety of ways and use them in different stages of our framework. A core aspect of our approach is leveraging generic resources to derive training pairs,
both natural and synthetic. On the natural front,  we use abundant data from a generic rewriting task: paraphrasing.
As for synthetic data, we leverage a general-purpose computational lexicon using its antonymy relation to generate polarity pairs.

In practice, we propose a framework that adopts a multi-step procedure which builds upon a general-purpose pre-trained sequence-to-sequence (seq2seq) model. 
First, we strengthen the model's ability to rewrite by  conducting a second phase of pre-training on natural pairs derived 
from an existing collection of generic paraphrases, as well as on synthetic pairs created using a general-purpose lexical resource.
Second, through an iterative back-translation \citep{hoang-etal-2018-iterative} approach, we train two models, each in a transfer direction, so that they can provide each other with synthetically generated pairs on-the-fly.
Lastly, we use our best resulting model to generate static synthetic pairs, which are then used offline as parallel training data.

\paragraph{Contributions}
Using a large pre-trained seq2seq model (1) we achieve state-of-the-art results for the two most popular style transfer tasks without task-specific parallel data. 
We show that (2) generic resources can be leveraged to derive parallel data for additional model pre-training, which boosts performance substantially and that (3) an iterative back-translation setting where models in the two transfer directions are trained simultaneously is successful, especially if enriched with a reward strategy.
We also offer (4) a theoretical contribution over the nature of the two tasks: while they are usually treated as the same task, our results 
suggest that they could possibly be treated separately.\footnote{All code at \url{https://github.com/laihuiyuan/Generic-resources-for-TST}.}

\section{Related Work}
\label{sec:related}

Style transfer is most successful if task-specific parallel data is available, as in the case of formality transfer \citep{rao-tetreault-2018}. 
Like in most NLP tasks, large pre-trained models
have been shown to provide an excellent base for fine-tuning in a supervised setting
\citep{chawla--semi, lai-etal-2021}.

Since parallel data for fine-tuning such large models for style transfer is scarce, a substantial amount of work has gone into methods for creating artificial sentence pairs so that models can be trained in a supervised regime. 

One way to do this is to artificially generate parallel data via back-translation, so that training pairs are created on-the-fly during the training process itself \citep{Zhirui-2018, lample2019multipleattribute, prabhumoye-2018, fuli-2019}. In these systems, one direction’s outputs and its inputs can be used as pairs to train the model of the opposite transfer direction. 

Another common strategy is to use style-word-editing \citep{li-etal-2018-delete, xu-etal-2018-unpaired, wu-etal-2019-hierarchical-reinforced, lee-2020-stable} to explicitly separate content and style.
These approaches first detect relevant words in the source and then do operations like deleting, inserting and combining to create the pair's target. Back-transferring is generally used to reconstruct the source sentence for training, so that pairs are also made on-the-fly.

\citet{lample2019multipleattribute} provide evidence that disentangling style and content to learn  distinct representations 
 \citep{Shen2017, Fu2018StyleTI, john-2019, xiaoyuan-ijcai}  is not necessary. 
 Reconstructing the source, instead, appears beneficial: it is used by \citet{dai-etal-2019-style} who pre-train a model on style transfer data with the Transformer architecture \citep{NIPS2017_3f5ee243}; and by \citet{zhou-etal-2020-exploring}, who 
 use an attentional seq2seq model that pre-trains the model to reconstruct the source sentence and re-predict its word-level style relevance.

\citet{fuli-2019} pre-train a LSTM-based seq2seq model \citep{bahdanau2014neural} using sentence pairs generated by a template-based baseline.
More recently, \citet{Jingjing-2020} proposed a two-stage strategy of search and learning for formality transfer where they perform a simulated annealing search \citep{liu-etal-2020-unsupervised} to obtain output sentences as pseudo-references, and then fine-tune GPT-2 \citep{radford-2019} with the resulting pairs.

The methods above create task-specific artificial pairs, some using pre-crafted manual rules or templates. We aim to overcome this by exploiting generic resources. 
Additionally, it is not evident which strategy works best for creating parallel data, whether offline or on-the-fly, and the simultaneous advantage of both strategies has not been fully explored. 
Lastly, \citet{chawla--semi} develop a semi-supervised model based on sequence-to-sequence pre-trained model (BART, \citet{lewis-etal-2020-bart}) using parallel training data and large amounts of non-parallel data, which achieves a significant performance. 
In previous work, we have also shown that a sequence-to-sequence pre-trained model (BART) outperforms a language model (GPT-2) in content preservation and overall performance when task-specific parallel training data is available \citep{lai-etal-2021}. 

Therefore, we use BART as generic base model; we enrich it with iterative back-translation to create training pairs on-the-fly. We also explore the advantage of further pre-training by creating pairs through generic resources, as well as the benefits of a final training using generated pairs.

\begin{table*}[t]
\resizebox{\linewidth}{!}{%
\centering
\small
\begin{tabular}{l|l|l|l}
\toprule
 \textsc{Dataset} & \textsc{Style} & \multicolumn{2}{c}{\textsc{Sentence-pair}}\\
 \midrule
 \multirow{2}{*}{GYAFC} & Informal & that is just my gut feeling. & no different between ages if the mind is near to eachother \\
 &  Formal & That is my personal opinion. & There is no difference between ages if the intellect is similar.\\
 \midrule
 \multirow{2}{*}{YELP} & Negative & this branch is getting worse and worse. & bad service in these areas and really ruined our visit. \\
 & Positive & this branch is getting better and better. & good service in these areas and really made our visit.\\
 \midrule
 \midrule
 \multirow{2}{*}{PARABANK 2} & Source & The bank is coming up on your left. & I guess I've always been pretty good with words.\\
 & Target & You have the bank on the left side. & I think narrating has always been my strong suit.\\
\bottomrule
\end{tabular}}
\caption{\label{tab:example}
Samples of each dataset.
}
\end{table*}


\begin{table}[t]
\resizebox{\linewidth}{!}{%
\centering
\small
\begin{tabular}{l|l|ccc|cc}
\toprule[1.2pt]
 \multirow{2}{*}{\textsc{Dataset}} & \multirow{2}{*}{\textsc{Style}} & \multicolumn{3}{c|}{\textsc{Paired}} & \multicolumn{2}{c}{\textsc{Unpaired}} \\
\cline{3-7} 
 & & Train & Valid & Test & Train & Valid\\
 \midrule
 \multirow{2}{*}{GYAFC [F\&R]} & Informal & 51,967 & 2,788 & 1,332 & N/A & N/A\\
 & Formal & 51.967 & 2,247 & 1,019 & N/A & N/A\\
 \midrule
 \multirow{2}{*}{YELP} & Negative & N/A & N/A & 500 & 177,218 & 2,000\\
 & Positive & N/A & N/A & 500 & 266,041 & 2,000\\
 \midrule
 \multirow{2}{*}{PARABANK 2} & Source & 1,132,289 & N/A & N/A & N/A & N/A\\
 & Target & 1,132,289 & N/A & N/A & N/A & N/A\\
\bottomrule[1.2pt]
\end{tabular}}
\caption{\label{tab:data-statistics}
Dataset Statistics.
}
\end{table}

\section{Tasks, Datasets, and Evaluation}
\label{sec:descriptions}

\subsection{Tasks and Datasets}\label{s:tasks_and_datasets}
The task of style transfer  is generally defined as the conversion of a text written in a given style to approximately the same text in a different style: style should be changed while preserving the original ``content''. We focus on the two most popular tasks, namely \textit{formality transfer} and \textit{polarity swap}, 
and use the two standard available datasets. Example pairs are shown in Table~\ref{tab:example}; statistics are in Table~\ref{tab:data-statistics}.

\paragraph{Formality Transfer Dataset}
Grammarly’s Yahoo Answers Formality Corpus (GYAFC) \cite{rao-tetreault-2018} is a dataset containing aligned formal and informal sentences from two domains: Entertainment \& Music (E\&M) and Family \& Relationships (F\&R). Parallel pairs are provided for training, validation, and test, with four human references for every test sentence. 
In the experiments we report in this paper we use data from the F\&R domain, which is the one more commonly used. 

\paragraph{Polarity Swap Dataset}
YELP is a dataset of business reviews on Yelp (with scores 1--5) processed by \citet{li-etal-2018-delete}.
Samples with a score greater than 3 are considered as positive  otherwise they are negative. The dataset comes in the form of large amounts of non-parallel data for training and development, while parallel pairs are provided for evaluation. 
For each test sentence, \citet{li-etal-2018-delete} provide one human reference; three additional human references are released by \citet{fuli-2019}.

\bigskip

\noindent Although these two tasks have been conflated in previous work as ``style transfer'', they are not exactly the same, which we hypothesise affects both their modelling and evaluation.
More specifically, in polarity swap the actual content is not exactly preserved (the message is actually the opposite), rather it's the general ``theme/topic'' that needs to be preserved. In formality transfer, instead, the ``translation'' happens really more at style level, and content needs to stay the same. This is evident if we look at examples in Table~\ref{tab:example} (top two blocks). In YELP, we can see that the theme-related words are expected to stay while changing the polarity words. Therefore, although the two sentences refer to the same event/concept, they convey opposite meanings. On the contrary, in formality transfer, an informal text should be  changed into a formal one, but the overall meaning should be preserved. In this sense, formality transfer can be seen much more as rewriting than polarity swap and can be conceived akin to the more general task of paraphrasing.

Leveraging this observation, we explore if 
paraphrase pairs can be used to make the model learn the basic task of ``rewriting'' in a first stage.
The advantage of using paraphrases is the large amount of parallel data available. 
Specifically, we use PARABANK~2, a large-scale, diverse, collection of paraphrases \citep{hu-etal-2019-large}. 
Given the different nature of the two tasks, we expect this strategy to help more formality transfer than polarity swap, since the latter is much less of a rewriting task than the former. 
In spite of the differences highlighted above, we approach both tasks within the same framework for two reasons: (i) to compare to previous works, which have treated the tasks as manifestations of the same ``style transfer'' task; but also (ii) to observe if and how the tasks respond differently to modelling and evaluation metrics.

\subsection{Task Evaluation}
\label{subsec:taskeval}
The performance of text style transfer is commonly assessed on style strength and content preservation. For style strength, using a pre-trained style classifier is the most popular automatic evaluation strategy. For content preservation, $n$-gram-based matching metrics such as BLEU~\citep{papineni-etal-2002-bleu} are most commonly used. However, these metrics usually fail to recognise  information beyond the lexical level. Since word embeddings \citep{Mikolov-2013, pennington-etal-2014-glove} have become the prime alternative to $n$-gram-based matching to capture similarity, embeddings-based metrics have also been developed \citep{Fu2018StyleTI}. However, embedding-based metrics like cosine similarity still work at the token-level, and might fail to capture the overall semantics of a sentence. 

To overcome such limitations, recent work has developed \textit{learnable metrics}, which attempt to directly optimize correlation with human judgments. These metrics, with the prime examples of BLEURT \citep{sellam-etal-2020-bleurt} and COMET \citep{ rei-etal-2020-comet}, have recently shown promising results in machine translation evaluation. To the best of our knowledge, only our previous work used BLEURT in the evaluation of formality style transfer models~\citep{lai-etal-2021}; we are now proposing to use it also for the evaluation of polarity swap, and to add COMET to the pool of evaluation metrics to be 
systematically adopted in the evaluation of text style transfer tasks. 

Therefore, in addition to BLEU, which allows us to compare to previous work, we also use BLEURT and COMET. Let us bear in mind that ``content preservation'' does not mean exactly the same thing for the two tasks that we consider (cf. Section~\ref{s:tasks_and_datasets}), so that we might observe different reactions to different evaluation measures for the two tasks.

\begin{figure*}[t]
    \centering
    \includegraphics[scale=0.69]{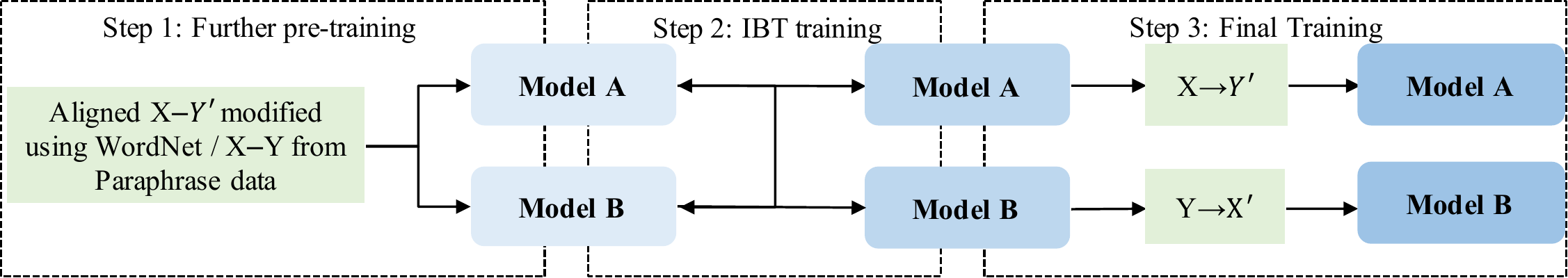}
    \caption{General overview of our  pipeline.}
    \label{fig:pipline-overview}
\end{figure*}

\section{Approach}

We propose a framework that adopts a multi-step procedure on top of the large pre-trained seq2seq model BART~\citep{lewis-etal-2020-bart}. 

Given a source sentence $\bm{x}=\{x_{1}, \cdots, x_{n}\}$ of length $n$ with style $s_{1}$, the goal of text style transfer is to generate a sentence $\bm{y}$
with style $s_{2}$, preserving the source sentence's meaning in formality transfer or the source sentence's theme in polarity swap.\footnote{In what follows, we use the term ``content'' in a more general way to refer to both cases.}
Formally, 
the objective is to minimize the following negative log likelihood:

\begin{equation}
\label{eq:bart-loss}
    L_(\phi)=
    -\Sigma_{i}\textrm{log}(p(y_{i}|y_{1:i-1}, \bm{x};\phi))
\end{equation}

\noindent where $\phi$ are the parameters of BART.

Our framework can be conceived as a pipeline, visualised in 
Figure~\ref{fig:pipline-overview}. At the core of the framework are two BART models (model A and model B), one for each transfer direction.
Since the main challenge in unpaired style transfer is that we cannot directly employ supervision (i.e. task-specific parallel training pairs), we explore and evaluate different ways of creating and using sentence pairs at different stages of the pipeline.

First, we strengthen the model's ability to rewrite by conducting a second phase of pre-training on natural pairs derived from an existing collection of generic paraphrases, as well as on synthetic pairs created using a general-purpose lexical resource (\textit{Step~1}, Section~\ref{subsec:fur-pre}).

Second, we use iterative back-translation  with several reward strategies to train the two models in both transfer directions simultaneously; sentence pairs are created on the fly (\textit{Step~2}, Section~\ref{subsec:IBT}). 

Third, we create high-quality synthetic pairs using our best systems from the previous step, to create a static resource of parallel data that can be used to train new transfer models (\textit{Step~3}, Section~\ref{subsec:high-qua}).

\subsection{Further Pre-training: Learning to Rewrite}
\label{subsec:fur-pre}

As hinted at in Section~\ref{sec:descriptions}, style transfer can be seen as a specific way of \textit{paraphrasing}. On the basis of this intuition,  
we hypothesise that generic paraphrase data, which already exists in much larger amounts than task-specific style transfer data, can be useful for text style transfer in terms of teaching the models the more generic task of ``rewriting''.
For polarity swap, which is less of a rewriting task than formality transfer, as the meaning is reversed rather than preserved, we also create synthetic pairs using a general-purpose lexical resource.

Using the natural and the synthetic pairs we conduct a second phase of pre-training. We expect this strategy to help specifically with  
content preservation, which is known to be the most difficult part of style transfer, especially in an unsupervised setting~\citep{Abhilasha-2020,lai-etal-2021}.

\begin{figure*}[t]
    \centering
    \includegraphics[scale=0.7]{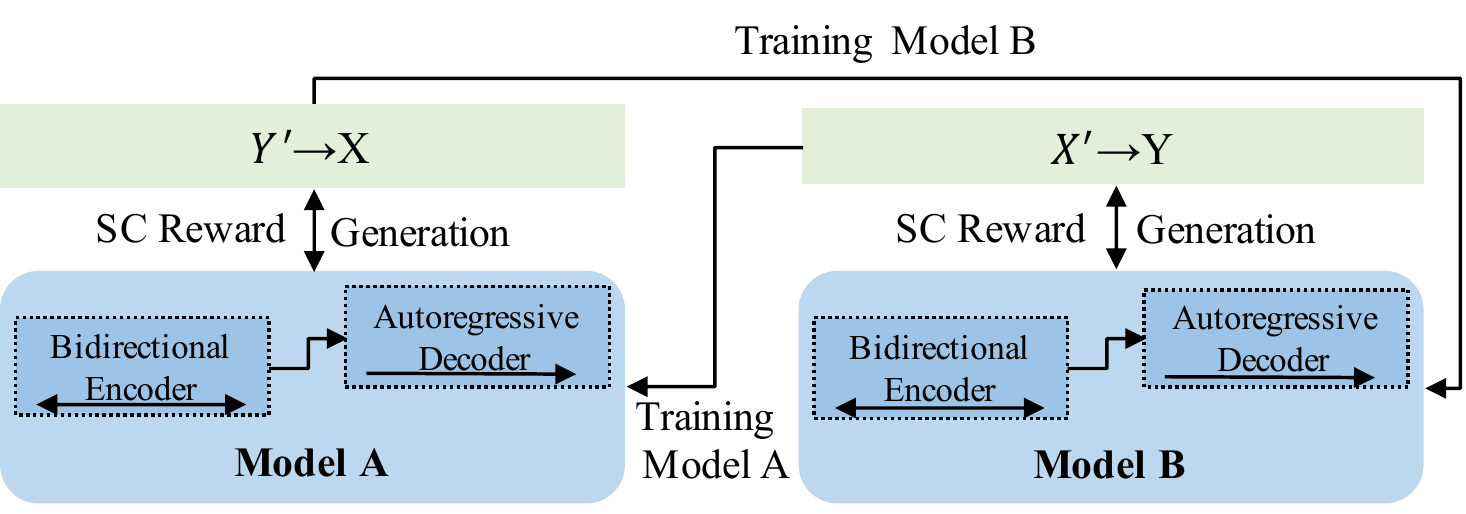}
    \caption{General overview of IBT training.}
    \label{fig:ibt-overview}
\end{figure*}

\paragraph{Generic Training Pairs} We use data from PARABANK~2 to make the model learn the basic task of “rewriting''. We use this dataset in its entirety or filtered (models M1.1 and M1.2 in Table~\ref{tab:ablation-study}). In the first case, the whole of the paraphrase pairs from PARABANK~2 are used to further pre-train the model.
In the second case, we follow the rationale that not all pairs are equally relevant for our tasks, and selecting task-specific ones could be beneficial. For instance, while both PARABANK~2 pairs in Table~\ref{tab:example} are good examples of rewriting, the one on the right is more meaningful in terms of formality transfer. Therefore, we train two binary style classifiers, one for formality and one for polarity, using TextCNN \citep{kim-2014} on the training sets of GYAFC and YELP. These classifiers are then used to automatically select more strongly style-opposed pairs.
The resulting filtered paraphrase subset $\bm{D}_{p}$ is such a set of pairs:

\begin{equation}
\label{reward-filter}
    \bm{D}_{p}=\{(\bm{x},\bm{y})|(p(s_{1}|\bm{x}) + p(s_{2}|\bm{y}))/2>\sigma\}
\end{equation}

\noindent where $p(s_{i}|*)$ is the probability of a sentence being a style $s_i$, predicted by the style classifier, and $\sigma$ is the threshold  for data selection\footnote{$\sigma$ = 0.85 in our experiments.}; $\bm{x}$ and $\bm{y}$ constitute the sentence pair. 

\paragraph{Synthetic Pairs for Polarity Swap}
Due to the nature of polarity swap, we expect that 
even filtered paraphrases might not benefit polarity swap as much as formality. 
We therefore add another strategy to enhance polarity swap rewriting and create pairs for further pre-training exploiting a general-purpose lexical resource (model M1.3 in Table~\ref{tab:ablation-study}). Specifically, we use SentiWordNet~\citep{Baccianella-2010} to obtain words' sentiment scores to detect the polarity of each word in the sentence. 
To maximise the quality of synthetic pairs, we select sentences that contain one polarity word only, and swap that one with its WordNet antonym \citep{Miller-wordnet}.
The new synthetic sentence is regarded as the target sentence corresponding to the original sentence.

\medskip

\noindent The generic/filtered/synthetic pairs are used for a second phase of seq2seq pre-training for BART. Examples of these pairs are  in Appendix~\ref{append:sample-para}.

\subsection{Iterative Back-translation and Rewards: Pairs on-the-fly}
\label{subsec:IBT}
After further pre-training BART, we use iterative back-translation to train two models, each in a transfer direction, so that they can provide one another with synthetically generated pairs on-the-fly. We obtain pseudo-parallel data via back-transfer: the outputs of one direction are used to provide the supervision to train the model of the opposite direction (Figure~\ref{fig:ibt-overview}).
To explicitly guide the model to preserve the content and to apply the target style, we add content and style rewards in a reinforcement learning fashion (models M2.* in Table~\ref{tab:ablation-study}).

\paragraph{Rewarding Style Strength} To provide a explicit signal to teach the model to change the sentence's style, a style classifier (SC) based reward is used to 
push the model to change the sentence into the target style.
For this SC reward, which evaluates how well the transferred sentence $\bm{y}'$ matches the target style, we reuse the style classifier trained for selecting paraphrase data (Section~\ref{subsec:fur-pre}). The SC's confidence in each transfer direction is  

\begin{equation}
\label{cls-softmax}
    p(s_{i}|\bm{y}') = softmax_{i}(TextCNN(\bm{y}', \theta))
\end{equation}
where $i$ = \{1,2\} and $\theta$ are the parameters of the style classifier, fixed during training transfer models. Formally, the reward is

\begin{equation}
\label{eq:reward-cls}
    R_{sc}=\lambda_{sc}[p(s_{2}|\bm{y}') - p(s_{1}|\bm{y}')]
\end{equation}
where $s_{1}$ and $s_{2}$ are source style and target style, respectively. $\bm{y}'$ is the generated target sentence sampled from the distribution of model outputs at each decoding time step. 

We apply the SC reward in two ways: in the supervised training process using pseudo-parallel data (SC0); and in the process of \textit{generating} pseudo-parallel data itself (SC1). For the latter, we generate text in the target style by sampling the distribution of model outputs, while at the same time use the SC reward to feed back its corresponding style signals to the model. 

\paragraph{Rewarding Content Preservation}

Following \citet{Abhilasha-2020}, we use a BLEU-based reward, formulated as follows:

\begin{small}
\begin{equation}
\label{eq:reward-bleu}
    R_{\textrm{bleu}}=
    \lambda_{\textrm{bleu}}[BLEU(\bm{y}_{s_{i}}^{'}, \bm{x})-BLEU(\bm{y}_{s_{i}}^{s}, \bm{x})]
\end{equation}
\end{small}
where $\bm{y}_{s_{i}}^{s}$ 
is the generated sentence in target style $s_{i}$ sampled from the distribution of model outputs at each time step in decoding, and $\bm{y}_{s_{i}}^{'}$ is obtained by greedily maximizing the distribution.

Since new-generation metrics show promising results in evaluation (Section~\ref{sec:descriptions}), we use BLEURT also as an alternative metric to BLEU in the reward strategy, expecting it might be better at measuring semantics at the sentence level. Formally, we formulate the BLEURT-based reward as

\begin{equation}
\label{eq:reward-bleurt}
 R_{\textrm{bleurt}}=
    \lambda_{\textrm{bleurt}}[BLEURT(\bm{y}_{s_{i}}^{s}, \bm{x})]\\
\end{equation}
where $\bm{y}_{s_{i}}^{s}$ is the generated sentence in target style $s_{i}$ sampled from the distribution of model outputs.

\paragraph{Gradients and Objectives}

We use the policy gradient algorithm~\citep{Williams-1992} to maximize the expected reward 
of the generated sentence $\bm{y}^{s}$, whose gradient with respect to the parameters $\phi$ of the neural network model is estimated by sampling:

\begin{equation} 
\begin{split}
\label{eq:gradient}
\nabla_{\phi}J(\phi)&=E[R\cdot\nabla_{\phi}log(P(\bm{y}^{s}|\bm{x};\phi))]
\end{split}
\end{equation}
where $\nabla_{\phi}J(\cdot)$ is the gradient of objective function $J(\cdot)$ with respect to model parameters $\phi$, 
$E(\cdot)$ is the expectation, $R$ is the reward of the sequence $\bm{y}^{s}$ that is sampled from the distribution of model outputs at each decoding time step. The overall objectives are the combination of the base model's loss (Eq.~\ref{eq:bart-loss}) and the policy gradient of rewards (Eq.~\ref{eq:gradient}) which are used to train our framework end-to-end.

\subsection{Final Training: High-quality Pairs}
\label{subsec:high-qua}
As a final step, we let our best models generate pairs to create a static resource of parallel data. We feed the system source sentences randomly picked from the training sets and generate the corresponding sentences in the target style. We then select high-quality pairs using BLEURT and our style classifier. The resulting dataset $\bm{D}_{h}$ is  a set of pairs:

\begin{equation}
\begin{split}
\label{pairs-filter}
    \bm{D}_{h} &=\{(\bm{x},\bm{y^{'}})|BLEURT(\bm{x},\bm{y^{'}})>\sigma_{c}\\
    & \qquad and\ (p(s_{1}|\bm{x}) + p(s_{2}|\bm{y^{'}}))/2>\sigma_{s} \}
\end{split}
\end{equation}

\noindent where $\bm{x}$ and $\bm{y^{'}}$ are the source sentence and generated sentence, respectively. $p(s_{i}|*)$ is the probability of a sentence being of style $s_i$ as predicted by the style classifier, and $\sigma_{*}$ is the threshold for data selection regarding content and style.\footnote{$\sigma_{c}$ = 0.15 and $\sigma_{s}$ = 0.9 in our experiments.}  

Finally, these pairs are used to fine-tune the original BART with all reward strategies, so as to train new transfer models in a supervised way (model~M3.1 in Table~\ref{tab:ablation-study}).

\section{Experiments}

\begin{table*}[ht]
\centering
\resizebox{\linewidth}{!}{%
\begin{tabular}{l|ccccc|ccccc}
\toprule[2pt]
\multicolumn{1}{r|}{\sc Dataset~~~} & \multicolumn{5}{c|}{\textsc{GYAFC (Formality Transfer)}} & \multicolumn{5}{c}{\textsc{YELP (Polarity Swap)}}\\
\hline
 \textsc{Model} & BLEURT & COMET & BLEU & ACC & HM & BLEURT & COMET & BLEU & ACC & HM \\
 \toprule
M0: Original BART & -0.116 & 0.242 & 0.414 & 0.333 & 0.369 & -0.388 & -0.146 & 0.309 & 0.022 & 0.041\\ \hline
\rowcolor{LightGray}
 \multicolumn{11}{c}{STEP~1: Further pre-training}\\ \hline
M1.1: Further pre-training using whole dataset & 0.012 & 0.209 & 0.420 & 0.357 & 0.386 & -0.412 & -0.282 & 0.179 & 0.040 & 0.065\\
M1.2: Further pre-training  using subset & 0.011 & 0.225 & 0.441 & 0.693 & 0.539 & -0.347 & -0.178 & 0.247 & 0.166 & 0.199\\
M1.3:  Further pre-training  using synthetic data & - & - & - & - & - & -0.321 & -0.074 & 0.326 & 0.189 & 0.239\\
 \hline
 \rowcolor{LightGray}
 \multicolumn{11}{c}{STEP~2: IBT + Rewards}\\ \hline
M2.1: IBT + all rewards with M0 & -0.010 & 0.292 & 0.507 & 0.836 & 0.631 & -0.229 & -0.017 & 0.298 & 0.826 & 0.438\\
M2.2: IBT + all rewards with M1.2 &\textbf{0.041} & 0.318 & 0.553 & \textbf{0.932} & \textbf{0.694} & -0.176 & 0.026 & 0.295 & 0.853 & 0.438\\
M2.3: IBT + all rewards with M1.3 &
- & - & - & - & - & -0.246 & -0.035 & 0.302 & 0.884 & 0.450\\
 \midrule
M2.4: M2.2 except BLEURT & 0.033 & 0.313 & 0.552 & 0.929 & 0.693 & -0.187 & 0.001 & 0.285 & 0.860 & 0.428\\
M2.5: M2.2 except BLEU & \textbf{0.041} & 0.320 & 0.551 & 0.925 & 0.691 & \textbf{-0.149} & 0.031 & 0.295 & 0.784 & 0.429 \\
M2.6: M2.2 except SC0 & 0.024 & \textbf{0.321} & 0.544 & 0.928 & 0.686 & -0.195 & -0.016 & 0.286 & 0.881 & 0.432\\
M2.7: M2.2 except SC1 & 0.039 & 0.318 & 0.555 & 0.873 & 0.679 & -0.176 & 0.039 & \textbf{0.331} & 0.500 & 0.398\\
 \hline
 \rowcolor{LightGray}
 \multicolumn{11}{c}{STEP~3: Offline training (Model used: original BART + Rewards)}\\ \hline
M3.1: training pairs generated with M2.2 (GYAFC) /   M2.3 (YELP)
& 0.030 & \textbf{0.321} & \textbf{0.560} & 0.904 & 0.692 & -0.183 & \textbf{0.046} & 0.316 & \textbf{0.887} & \textbf{0.466}\\
M3.2: training pairs are subset of paraphrase data (same as in M1.2) & 0.012 & 0.229 & 0.455 & 0.783 & 0.576 & -0.338 & -0.221 & 0.215 & 0.457 & 0.292\\
\bottomrule[2pt]
\end{tabular}}
\caption{\label{tab:ablation-study}
Results for the different steps of the pipeline. SC0 is the SC reward used in the supervised training process using pseudo-parallel data, and SC1 is used in the process of generating pseudo-parallel data.
}
\end{table*}

\begin{table*}[ht]
\resizebox{\linewidth}{!}{%
\centering
\small
\begin{tabular}{l|l|l|cccc}
\toprule[1.5pt]
  \textsc{Tasks} & \textsc{Model} & \textsc{Sentence} & BLEU & BLEURT & COMET & ACC \\
  \hline
  \multirow{8}{*}{\makecell[c]{Informal $\rightarrow$ Formal}} & Source & So if you're set on that, that's the way to go!! & \multicolumn{4}{c}{-} \\
  \cline{2-7}
  & M0   & so if you're set on that, that's the way to go!! & 0.417 & 0.175 & 0.568 & 0.000\\
  & M1.1 & so if you want to do this, this is the way to go! & 0.301 & 0.204 & 0.354 & 0.003\\
  & M1.2 & If you want to do this, this is the way to go. & 0.416 & 0.339 & 0.433 & 0.855\\
  & M2.1 & So if you're set on that, that is the way to go. & 0.763 & 0.525 & 0.689 & 0.179\\
  & M2.2 & So, if you are set on that, then that is the way to go. & \textbf{0.884} & 0.456 & 0.722 & \textbf{0.880}\\
  & M3.1 & So if you are set on that, that is the way to go. & 0.541 & \textbf{0.941} & \textbf{0.734} & 0.617\\
  & M3.2 & If you're on board, that's the way to go. & 0.352 & 0.200 & 0.311 & 0.552\\
\toprule[1pt]  
  \multirow{10}{*}{\makecell[c]{Positive $\rightarrow$ Negative}} & Source & the staff are all super friendly and on top of there jobs.       & \multicolumn{4}{c}{-}\\
  \cline{2-7}
  & M0   & the staff are all super friendly and on top of there jobs.        & 0.163 & -0.561 & -0.169 & 0.000\\
  & M1.1 & all the staff are very friendly and they're doing their jobs well.& 0.107 & -0.571 & -0.301 & 0.003\\
  & M1.2 & the staff are all super friendly and on top of each same jobs.    & 0.149 & -0.662 & -0.507 & 0.000\\
  & M1.3 & the staff are all super unfriendly and on top of there jobs.      & 0.151 & -0.239 & 0.095  & \textbf{1.000}\\
  & M2.1 & the staff are all super rude and on top of there jobs.            & 0.151 & -0.513 & 0.048  & \textbf{1.000}\\
  & M2.2 & the staff are all super rude and on top of there jobs.            & 0.151 & -0.513 & 0.048  & \textbf{1.000}\\
  & M2.3 & the staff are all super rude and on top of there jobs.            & 0.151 & -0.513 & 0.048  & \textbf{1.000}\\
  & M3.1 & the staff are not super friendly or on top of there jobs.         & \textbf{0.320} & \textbf{0.322}  & \textbf{0.621}  & \textbf{1.000}\\
  & M3.2 & the staff are so friendly and they're doing their jobs.           & 0.148 & -0.663 & -0.326 & 0.001\\
  
\bottomrule[1.5pt]
\end{tabular}}
\caption{\label{tab:example-output-step}
Example outputs for the different steps of the pipeline and their corresponding evaluation results. Note that ACC represents style confidence here.
}
\end{table*}

All experiments are implemented atop Huggingface Transformers \citep{wolf-etal-2020-transformers}, taking
the BART base model (139M parameters) for our experiments. We train our framework using the Adam optimiser \citep{diederik-kingma-2015} with the initial learning rate $1e^{-5}$. The batch size is set to 32. The final values $\lambda$ for style and content rewards are both set to 1 based on validation results. Both WordNet and SentiWordNet are used from NLTK \footnote{\url{https://www.nltk.org/}}.

\subsection{Evaluation Metrics}
To assess style and content we use common metrics for this task. For content preservation we add two learnable metrics, which we hope will be adopted from now on, to glean better insights into the systems' behaviour in the two tasks  (Section~\ref{subsec:taskeval}). 

We measure \textbf{style strength} automatically by evaluating the target style accuracy of transferred sentences. We use the style classifiers trained for  selecting paraphrase data (Section~\ref{subsec:fur-pre}).
The classifiers have an accuracy of 92.6\% and 98.1\% on the test sets of F\&R and YELP, respectively.

To assess \textbf{content preservation}, we follow previous work and calculate BLEU\footnote{We use \texttt{multi-bleu.perl} with default settings.} between the generated sentence and the human reference(s). Additionally, we compute BLEURT and COMET\footnote{COMET is designed to also take input sentences into account, but our evaluations including them yielded lower correlations with human judgements. This might be because in COMET training input and output are different languages.}.
As the human references for YELP are released from different researchers and appear to differ quite a lot in nature (see Appendix~\ref{append:sample-ref} for examples), we provide two evaluation results: one using the first human reference only (Table~\ref{tab:ablation-study}), and the other using all four (Appendix~\ref{append:results-four-ref}).

As \textbf{overall score}, for a  direct comparison to previous work \cite{fuli-2019, zhou-etal-2020-exploring,lai-etal-2021} we compute the harmonic mean (HM) of style accuracy and BLEU.

\subsection{Results}
\label{sec:results}

\noindent Table~\ref{tab:ablation-study} reports results for each step.\footnote{Results for more models per step are in Appendix~\ref{append:ablation-study}.}

Results of Step~1 show that using paraphrase data benefits more formality transfer than polarity swap, confirming the latter is much less of a rewriting task than the former. Filtering paraphrases to a subset closer to the task (M1.2) substantially helps formality and yields some improvement in polarity. WordNet-derived synthetic pairs (M1.3) are definitely a better strategy for polarity.\footnote{The WordNet-based strategy could in principle be used on its own to solve the polarity swap task with no learning involved, but results prove it insufficient: BLEURT: -0.475; COMET: -0.221; BLEU: 0.296; ACC: 0.206; HM: 0.243.}

The first block of Step~2 confirms that further pre-training significantly improves performance on formality transfer (compare M2.2 with M2.1). This results in the best model for formality transfer. For polarity, instead, we see improvement from further pre-training only when  using WordNet-based synthetic pairs (compare M2.2 with M2.3). Overall, in Step~2 we see that combining SC rewards and content-related rewards results in the best balance regarding content preservation and style strength.

In Step~3, we see that the model trained with high-quality synthetic pairs (M3.1) achieves the best overall performance on polarity swap. For comparison, we use the subset of paraphrase data as training pairs in place of the generated pairs, and see that performance is lower (M3.2). 

\begin{table*}[t]
\centering
\resizebox{\linewidth}{!}{%
\begin{tabular}{l|ccccc|l|ccccc}
\toprule[2pt]
 \multicolumn{6}{c|}{\textsc{GYAFC (Formality Transfer)}} & \multicolumn{6}{c}{\textsc{YELP (Polarity Swap)}}\\
 \hline
 \textsc{Model} & BLEURT & COMET & BLEU & ACC & HM & \textsc{Model} & BLEURT & COMET & BLEU & ACC & HM \\
\toprule
 Input Copy & -0.114 & 0.272 & 0.474 & 0.120 & 0.192 & Input Copy & -0.383 & -0.139 & 0.312 & 0.019 & 0.036\\
 UnsuperMT \citep{Zhirui-2018} & -0.665 & -0.446 & 0.327 & 0.670 & 0.439 & Style-Transformer \citep{dai-etal-2019-style} & -0.469 & -0.269 & 0.282 & 0.857 & 0.424\\
 DualRL \citep{fuli-2019} & -0.589 & -0.451 & 0.404 & 0.654 & 0.499 & DualRL \citep{fuli-2019} & -0.385 & -0.202 & 0.278 & 0.894 & 0.424\\
 StyIns \citep{xiaoyuan-ijcai} & -0.395 & -0.112 & 0.458 & 0.761 & 0.573 & StyIns \citep{xiaoyuan-ijcai} & -0.576 & -0.390 & 0.250 & \textbf{0.924} & 0.394\\
 Zhou's \citep{zhou-etal-2020-exploring} & -0.454 & -0.203 & 0.447 & 0.799 & 0.573 & Zhou's \citep{zhou-etal-2020-exploring} & -0.270 & -0.051 & 0.302 & 0.865 & 0.448\\
 *TGLS (\citet{Jingjing-2020}; 0 $\rightarrow$ 1) & - & - & 0.603 & - & - & DGST \citep{li-etal-2020-dgst} & -0.421 & -0.240 & 0.268 & 0.781 & 0.399\\
 \hline
 Ours (M2.2; lowercase) & \textbf{0.009} & \textbf{0.328} & \textbf{0.563} & \textbf{0.866} & \textbf{0.682} & Ours (M3.1) & \textbf{-0.183} & \textbf{0.046} & \textbf{0.316} & 0.887 & \textbf{0.466}\\
 Ours (M2.2; lowercase; 0 $\rightarrow$ 1) & - & - & \textbf{0.741} & - & - & - & \multicolumn{5}{c}{-}\\
\bottomrule[2pt]
\end{tabular}}
\caption{\label{tab:results}
Comparison with other systems. Notes: (i) we lowercase the GYAFC texts for a fairer comparison to previous works, as they do so; (ii) if the output of previous work is available, we re-calculate the scores using our  metrics. Otherwise we take the scores from the paper and mark this with a (*); (iii) we report our results on informal-to-formal (0 $\rightarrow$ 1) alone to compare with \citet{Jingjing-2020}, who only transfer in this direction.
}
\end{table*}

Table ~\ref{tab:example-output-step} shows  example outputs of each step and their  evaluation results.\footnote{References and more examples are in Appendix~\ref{append:example-outputs-step-all}.} It is interesting to see the impact of paraphrase-based pre-training: for formality, in M1.1 and M1.2, the phrase ``if you want to do this'' is used in place of ``if you're set on that''. This rewriting ability can also be observed on the polarity swap (``on top of there jobs'' $\rightarrow$ ``they're doing their jobs well''; note also that using paraphrases seems to prompt better writing: ``there'' $\rightarrow$ ``their'', M1.1/M3.2, though this is not consistent throughout the models). For formality, the quality of the output gradually improves in Step~2, with M2.2 achieving the best performance on BLEU and style confidence (M2.2); the model trained with high-quality synthetic pairs (M3.1) has the highest BLEURT and COMET. In M3.2, trained on paraphrase pairs, we find nice variability again (``if you're on board''). For polarity, M1.3 (using WordNet-based synthetic pairs), swaps a polarity word with its antonym (``friendly'' $\rightarrow$ ``unfriendly''). In Step~2, the models are indeed changing the polarity of the sentence; finally, the model trained with high-quality pairs (M3.1) nicely changes ``and'' into ``or'' to get the right semantics (though it loses the correct form ``their'') and is scored best. Further exploration of combining generic  and task-specific rewriting appears very promising for these tasks.

As an additional curiosity-driven qualitative assessment of the behaviour of our models, we probed the polarity swap models with \textit{neutral} sentences.\footnote{This was a suggestion of a reviewer, and we found indeed that this perspective could provide helpful insights in the models' behaviour to be further studied in future work.} As a first example, we use ``the earth revolves around the sun.'' as the source sentence, and observe that the models in both transfer directions generate the same sentences as the input. With as input the neutral sentence ``there is a grocery store near my house.'', the model which transforms negative sentences into positive ones generates ``there is a great grocery store near my house.'' while into the other direction it generates ``there is no grocery store near my house.'' It is worth mentioning that all the training data comes from business reviews on YELP, and the first example is clearly outside that domain. For the second example, closer to the domain of YELP, the transformation proposed by the model is rather reasonable in terms of obtaining a positive (``great grocery store'') or negative (``no grocery store'') output. It is left to future research to investigate what it should mean to transform a neutral sentence into a positive/negative one, and how such a test can help to better understand the models' behaviour and the task itself.

\paragraph{Comparison to other systems}
To put our results in perspective, we compare our best system (M2.2 for formality and M3.1 for polarity in Table~\ref{tab:ablation-study}) against the most recent and  best performing unpaired systems. 
For formality: UnsuperMT \citep{Zhirui-2018}; DualRL \citep{fuli-2019}; StyIns \citep{xiaoyuan-ijcai}; Zhou's \citep{zhou-etal-2020-exploring}; TGLS \citep{Jingjing-2020}. For polarity: Style-Transformer \citep{dai-etal-2019-style}; DualRL \citep{fuli-2019}; StyIns \citep{xiaoyuan-ijcai}; Zhou's \citep{zhou-etal-2020-exploring}; DGST \citep{li-etal-2020-dgst}.\footnote{See Section~\ref{sec:related} for details on these models.} We also add a simple baseline that just copies the input as output. 

As visible in Table~\ref{tab:results},
 our models achieve the best overall performance on both tasks.
For formality transfer, this is true in all evaluation metrics. For polarity swap, StyIns 
has the highest style accuracy, while our model is better  on all other metrics.\footnote{A sample comparison of outputs is in Appendix ~\ref{append:output-comparison}.}

\begin{table}[t]
\resizebox{\linewidth}{!}{%
\centering
\small
\begin{tabular}{l|c|ccc}
\toprule[1.3pt]
\multirow{2}{*}{\textsc{Tasks}} & \multirow{2}{*}{N} & BLEURT & COMET & BLEU\\
\cline{3-5}
& & COMET & BLEU & BLEURT\\
\hline
\multirow{2}{*}{Formality Transfer} & \multirow{2}{*}{21} & 0.980 & 0.775 & 0.761\\
& & (p\textless0.01) & (p\textless0.01) & (p\textless0.01)\\
\hline
\multirow{2}{*}{Polarity Swap} & \multirow{2}{*}{21} & 0.968 & 0.671 & 0.479\\
& & (p\textless0.01) & (p\textless0.01) & (p=0.03)\\

\bottomrule[1.3pt]
\end{tabular}}
\caption{\label{tab:corr-auto}
Pearson correlation between evaluation metrics for content preservation over $N$ systems.
}
\end{table}

\subsection{Reflections on Tasks and Evaluation}

The strategy of making the model learn the basic task of ``rewriting'' in a first stage clearly benefits more formality transfer than polarity swap. This is not surprising, since the latter is not simply ``rewriting a sentence in a different stlye''; rather, the task involves changing the meaning of a sentence to obtain its opposite polarity, and thus, broadly put, its meaning. The fact that polarity swap cannot be regarded as a ``style change'' task is also evident from evaluation. Rather than only using BLEU, we suggested to also use BLEURT and COMET, and this provides us with additional evidence. Specifically, from Table~\ref{tab:corr-auto} we observe that BLEU has a high correlation with BLEURT/COMET for formality transfer but not for polarity swap. 

To glean further insights into this difference, we leverage human judgments released by \citet{li-etal-2018-delete} for YELP and see how they correlate with the used metrics. We calculate system-level Pearson correlation between the automatic evaluations and human judgment. 

Results show that while COMET and BLEURT highly correlate with human judgments, BLEU does so to a lesser extent, suggesting this might be a less strong measure to assess the goodness of polarity swap.\footnote{
Pearson's $r=.922$ for BLEURT, $r=.941$ for COMET, and $r=.901$ for BLEU. All $p<.001, N=7$.} Intuitively, if a system does not change the polarity it may still have a high $n$-gram overlap (high BLEU) while new-generation metrics do not have this problem. For formality this limitation of BLEU is not much of an issue, since meaning is not altered. Nevertheless, we suggest that the evaluation of style transfer and related tasks should use learned metrics whenever possible.

\section{Conclusions}

We proposed an unpaired approach that adopts a multi-step procedure based on the general-purpose pre-trained seq2seq model BART. 

Achieving state-of-the-art results on the two most popular ``style transfer'' tasks, we have shown the benefit of further pre-training using data derived from generic resources as well as the advantage of back-translation, paired with rewards, especially towards content preservation. We have also seen how leveraging paraphrases can enhance both variability and naturalness in the generated text.

Through experimental settings as well as the introduction of BLEURT and COMET as metrics, we have also highlighted how the two tasks we addressed differ, and should probably not be conflated into a single ``style tranfer'' label. Indeed, we show that they benefit from partially different modelling, and react differently to evaluation metrics, both key aspects  to improve future modelling of these tasks.

\section*{Acknowledgments}

This work was partly funded by the China Scholarship Council (CSC). The anonymous EMNLP reviewers provided us with useful comments which contributed to improving this paper and its presentation, so we're grateful to them. We would also like to thank the Center for Information
Technology of the University of Groningen for their
support and for providing access to the Peregrine
high performance computing cluster.

\section*{Ethics Statement}

All work that automatically generates and/or alters natural text could unfortunately be used maliciously. While we cannot fully prevent such uses once our models are made public, we do hope that writing about risks explicitly and also raising awareness of this possibility in the general public are ways to contain the effects of potential harmful uses. We are open to any discussion and suggestions to minimise such risks.

\bibliography{anthology,custom}

\begin{thebibliography}{38}
\expandafter\ifx\csname natexlab\endcsname\relax\def\natexlab#1{#1}\fi

\bibitem[{Baccianella et~al.(2010)Baccianella, Esuli, and
  Sebastiani}]{Baccianella-2010}
Stefano Baccianella, Andrea Esuli, and Fabrizio Sebastiani. 2010.
\newblock Sentiwordnet 3.0: An enhanced lexical resource for sentiment analysis
  and opinion mining.
\newblock volume~10.

\bibitem[{Bahdanau et~al.(2015)Bahdanau, Cho, and Bengio}]{bahdanau2014neural}
Dzmitry Bahdanau, Kyunghyun Cho, and Yoshua Bengio. 2015.
\newblock \href {http://arxiv.org/abs/1409.0473} {Neural machine translation by
  jointly learning to align and translate}.
\newblock \emph{arXiv preprint, arXiv: 1409.0473}.

\bibitem[{Chawla and Yang(2020)}]{chawla--semi}
Kunal Chawla and Diyi Yang. 2020.
\newblock Semi-supervised formality style transfer using language model
  discriminator and mutual information maximization.
\newblock In \emph{Findings of the Association for Computational Linguistics:
  EMNLP 2020}, pages 2340--2354.

\bibitem[{Dai et~al.(2019)Dai, Liang, Qiu, and Huang}]{dai-etal-2019-style}
Ning Dai, Jianze Liang, Xipeng Qiu, and Xuanjing Huang. 2019.
\newblock \href {https://doi.org/10.18653/v1/P19-1601} {Style transformer:
  Unpaired text style transfer without disentangled latent representation}.
\newblock In \emph{Proceedings of the 57th Annual Meeting of the Association
  for Computational Linguistics}, pages 5997--6007, Florence, Italy.
  Association for Computational Linguistics.

\bibitem[{Fu et~al.(2018)Fu, Tan, Peng, Zhao, and Yan}]{Fu2018StyleTI}
Zhenxin Fu, Xiaoye Tan, Nanyun Peng, Dongyan Zhao, and Rui Yan. 2018.
\newblock Style transfer in text: Exploration and evaluation.
\newblock In \emph{Proceedings of the 28th International Joint Conference on
  Artificial Intelligence}, pages 663--670.

\bibitem[{Hoang et~al.(2018)Hoang, Koehn, Haffari, and
  Cohn}]{hoang-etal-2018-iterative}
Vu~Cong~Duy Hoang, Philipp Koehn, Gholamreza Haffari, and Trevor Cohn. 2018.
\newblock \href {https://doi.org/10.18653/v1/W18-2703} {Iterative
  back-translation for neural machine translation}.
\newblock In \emph{Proceedings of the 2nd Workshop on Neural Machine
  Translation and Generation}, pages 18--24, Melbourne, Australia. Association
  for Computational Linguistics.

\bibitem[{Hu et~al.(2019)Hu, Singh, Holzenberger, Post, and
  Van~Durme}]{hu-etal-2019-large}
J.~Edward Hu, Abhinav Singh, Nils Holzenberger, Matt Post, and Benjamin
  Van~Durme. 2019.
\newblock Large-scale, diverse, paraphrastic bitexts via sampling and
  clustering.
\newblock In \emph{Proceedings of the 23rd Conference on Computational Natural
  Language Learning (CoNLL)}, pages 44--54.

\bibitem[{John et~al.(2019)John, Mou, Bahuleyan, and Vechtomova}]{john-2019}
Vineet John, Lili Mou, Hareesh Bahuleyan, and Olga Vechtomova. 2019.
\newblock Disentangled representation learning for non-parallel text style
  transfer.
\newblock In \emph{Proceedings of the 57th Annual Meeting of the Association
  for Computational Linguistics}, pages 424--434.

\bibitem[{Kim(2014)}]{kim-2014}
Yoon Kim. 2014.
\newblock Convolutional neural networks for sentence classification.
\newblock In \emph{Proceedings of the 2014 Conference on Empirical Methods in
  Natural Language Processing ({EMNLP})}, pages 1746--1751.

\bibitem[{Kingma and Ba(2015)}]{diederik-kingma-2015}
Diederik~P Kingma and Jimmy Ba. 2015.
\newblock \href {http://arxiv.org/abs/1412.6980} {Adam: A method for stochastic
  optimization}.
\newblock \emph{arXiv preprint, arXiv: 1412.6980}.

\bibitem[{Lai et~al.(2021)Lai, Toral, and Nissim}]{lai-etal-2021}
Huiyuan Lai, Antonio Toral, and Malvina Nissim. 2021.
\newblock \href {https://doi.org/10.18653/v1/2021.acl-short.62} {Thank you
  {BART}! rewarding pre-trained models improves formality style transfer}.
\newblock In \emph{Proceedings of the 59th Annual Meeting of the Association
  for Computational Linguistics and the 11th International Joint Conference on
  Natural Language Processing (Volume 2: Short Papers)}, pages 484--494,
  Online. Association for Computational Linguistics.

\bibitem[{Lample et~al.(2019)Lample, Subramanian, Smith, Denoyer, Ranzato, and
  Boureau}]{lample2019multipleattribute}
Guillaume Lample, Sandeep Subramanian, Eric Smith, Ludovic Denoyer,
  Marc'Aurelio Ranzato, and Y-Lan Boureau. 2019.
\newblock Multiple-attribute text rewriting.
\newblock In \emph{International Conference on Learning Representations}.

\bibitem[{Lee(2020)}]{lee-2020-stable}
Joosung Lee. 2020.
\newblock \href {https://www.aclweb.org/anthology/2020.inlg-1.25} {Stable style
  transformer: Delete and generate approach with encoder-decoder for text style
  transfer}.
\newblock In \emph{Proceedings of the 13th International Conference on Natural
  Language Generation}, pages 195--204, Dublin, Ireland. Association for
  Computational Linguistics.

\bibitem[{Lewis et~al.(2020)Lewis, Liu, Goyal, Ghazvininejad, Mohamed, Levy,
  Stoyanov, and Zettlemoyer}]{lewis-etal-2020-bart}
Mike Lewis, Yinhan Liu, Naman Goyal, Marjan Ghazvininejad, Abdelrahman Mohamed,
  Omer Levy, Veselin Stoyanov, and Luke Zettlemoyer. 2020.
\newblock {BART}: Denoising sequence-to-sequence pre-training for natural
  language generation, translation, and comprehension.
\newblock In \emph{Proceedings of the 58th Annual Meeting of the Association
  for Computational Linguistics}, pages 7871--7880.

\bibitem[{Li et~al.(2020{\natexlab{a}})Li, Li, Mou, Jiang, Lyu, and
  King}]{Jingjing-2020}
Jingjing Li, Zichao Li, Lili Mou, Xin Jiang, Michael~R. Lyu, and Irwin King.
  2020{\natexlab{a}}.
\newblock Unsupervised text generation by learning from search.
\newblock In \emph{Advances in Neural Information Processing Systems 33: Annual
  Conference on Neural Information Processing Systems 2020, NeurIPS 2020,
  December 6-12, 2020, virtual}.

\bibitem[{Li et~al.(2018)Li, Jia, He, and Liang}]{li-etal-2018-delete}
Juncen Li, Robin Jia, He~He, and Percy Liang. 2018.
\newblock \href {https://doi.org/10.18653/v1/N18-1169} {Delete, retrieve,
  generate: a simple approach to sentiment and style transfer}.
\newblock In \emph{Proceedings of the 2018 Conference of the North {A}merican
  Chapter of the Association for Computational Linguistics: Human Language
  Technologies, Volume 1 (Long Papers)}, pages 1865--1874, New Orleans,
  Louisiana. Association for Computational Linguistics.

\bibitem[{Li et~al.(2020{\natexlab{b}})Li, Chen, Lin, and
  Li}]{li-etal-2020-dgst}
Xiao Li, Guanyi Chen, Chenghua Lin, and Ruizhe Li. 2020{\natexlab{b}}.
\newblock {DGST}: a dual-generator network for text style transfer.
\newblock In \emph{Proceedings of the 2020 Conference on Empirical Methods in
  Natural Language Processing (EMNLP)}, pages 7131--7136.

\bibitem[{Liu et~al.(2020)Liu, Mou, Meng, Zhou, Zhou, and
  Song}]{liu-etal-2020-unsupervised}
Xianggen Liu, Lili Mou, Fandong Meng, Hao Zhou, Jie Zhou, and Sen Song. 2020.
\newblock Unsupervised paraphrasing by simulated annealing.
\newblock In \emph{Proceedings of the 58th Annual Meeting of the Association
  for Computational Linguistics}, pages 302--312.

\bibitem[{Luo et~al.(2019)Luo, Li, Zhou, Yang, Chang, Sui, and Sun}]{fuli-2019}
Fuli Luo, Peng Li, Jie Zhou, Pengcheng Yang, Baobao Chang, Zhifang Sui, and
  Xu~Sun. 2019.
\newblock A dual reinforcement learning framework for unsupervised text style
  transfer.
\newblock In \emph{Proceedings of the 28th International Joint Conference on
  Artificial Intelligence}, pages 5116--5122.

\bibitem[{Mikolov et~al.(2013)Mikolov, Sutskever, Chen, Corrado, and
  Dean}]{Mikolov-2013}
Tomas Mikolov, Ilya Sutskever, Kai Chen, Greg~S Corrado, and Jeff Dean. 2013.
\newblock \href
  {https://proceedings.neurips.cc/paper/2013/file/9aa42b31882ec039965f3c4923ce901b-Paper.pdf}
  {Distributed representations of words and phrases and their
  compositionality}.
\newblock In \emph{Advances in Neural Information Processing Systems},
  volume~26, page 3111–3119.

\bibitem[{Miller(1995)}]{Miller-wordnet}
George~A. Miller. 1995.
\newblock \href {https://doi.org/10.1145/219717.219748} {Wordnet: A lexical
  database for english}.
\newblock \emph{Commun. ACM}, 38(11):39–41.

\bibitem[{Papineni et~al.(2002)Papineni, Roukos, Ward, and
  Zhu}]{papineni-etal-2002-bleu}
Kishore Papineni, Salim Roukos, Todd Ward, and Wei-Jing Zhu. 2002.
\newblock \href {https://doi.org/10.3115/1073083.1073135} {{B}leu: a method for
  automatic evaluation of machine translation}.
\newblock In \emph{Proceedings of the 40th Annual Meeting of the Association
  for Computational Linguistics}, pages 311--318, Philadelphia, Pennsylvania,
  USA. Association for Computational Linguistics.

\bibitem[{Pennington et~al.(2014)Pennington, Socher, and
  Manning}]{pennington-etal-2014-glove}
Jeffrey Pennington, Richard Socher, and Christopher Manning. 2014.
\newblock \href {https://doi.org/10.3115/v1/D14-1162} {{G}love: Global vectors
  for word representation}.
\newblock In \emph{Proceedings of the 2014 Conference on Empirical Methods in
  Natural Language Processing ({EMNLP})}, pages 1532--1543, Doha, Qatar.
  Association for Computational Linguistics.

\bibitem[{Prabhumoye et~al.(2018)Prabhumoye, Tsvetkov, Salakhutdinov, and
  Black}]{prabhumoye-2018}
Shrimai Prabhumoye, Yulia Tsvetkov, Ruslan Salakhutdinov, and Alan~W Black.
  2018.
\newblock Style transfer through back-translation.
\newblock In \emph{Proceedings of the 56th Annual Meeting of the Association
  for Computational Linguistics (Volume 1: Long Papers)}, pages 866--876.

\bibitem[{Radford et~al.(2019)Radford, Wu, Child, Luan, Amodei, and
  Sutskever}]{radford-2019}
Alec Radford, Jeff Wu, Rewon Child, David Luan, Dario Amodei, and Ilya
  Sutskever. 2019.
\newblock Language models are unsupervised multitask learners.

\bibitem[{Rao and Tetreault(2018)}]{rao-tetreault-2018}
Sudha Rao and Joel Tetreault. 2018.
\newblock Dear sir or madam, may {I} introduce the {GYAFC} dataset: Corpus,
  benchmarks and metrics for formality style transfer.
\newblock In \emph{Proceedings of the 2018 Conference of the North {A}merican
  Chapter of the Association for Computational Linguistics: Human Language
  Technologies}, pages 129--140.

\bibitem[{Rei et~al.(2020)Rei, Stewart, Farinha, and
  Lavie}]{rei-etal-2020-comet}
Ricardo Rei, Craig Stewart, Ana~C Farinha, and Alon Lavie. 2020.
\newblock {COMET}: A neural framework for {MT} evaluation.
\newblock In \emph{Proceedings of the 2020 Conference on Empirical Methods in
  Natural Language Processing (EMNLP)}, pages 2685--2702, Online. Association
  for Computational Linguistics.

\bibitem[{Sancheti et~al.(2020)Sancheti, Krishna, Srinivasan, and
  Natarajan}]{Abhilasha-2020}
Abhilasha Sancheti, Kundan Krishna, Balaji~Vasan Srinivasan, and Anandhavelu
  Natarajan. 2020.
\newblock Reinforced rewards framework for text style transfer.
\newblock In \emph{Advances in Information Retrieval}, pages 545--560.

\bibitem[{Sellam et~al.(2020)Sellam, Das, and Parikh}]{sellam-etal-2020-bleurt}
Thibault Sellam, Dipanjan Das, and Ankur Parikh. 2020.
\newblock {BLEURT}: Learning robust metrics for text generation.
\newblock In \emph{Proceedings of the 58th Annual Meeting of the Association
  for Computational Linguistics}, pages 7881--7892.

\bibitem[{Shen et~al.(2017)Shen, Lei, Barzilay, and Jaakkola}]{Shen2017}
Tianxiao Shen, Tao Lei, Regina Barzilay, and Tommi Jaakkola. 2017.
\newblock Style transfer from non-parallel text by cross-alignment.
\newblock In \emph{Proceedings of the 31st International Conference on Neural
  Information Processing Systems}, pages 6833--6844.

\bibitem[{Vaswani et~al.(2017)Vaswani, Shazeer, Parmar, Uszkoreit, Jones,
  Gomez, Kaiser, and Polosukhin}]{NIPS2017_3f5ee243}
Ashish Vaswani, Noam Shazeer, Niki Parmar, Jakob Uszkoreit, Llion Jones,
  Aidan~N Gomez, \L~ukasz Kaiser, and Illia Polosukhin. 2017.
\newblock Attention is all you need.
\newblock In \emph{Advances in Neural Information Processing Systems},
  volume~30. Curran Associates, Inc.

\bibitem[{Williams(1992)}]{Williams-1992}
Ronald~J. Williams. 1992.
\newblock Simple statistical gradient-following algorithms for connectionist
  reinforcement learning.
\newblock \emph{Machine Learning}, 8:229–256.

\bibitem[{Wolf et~al.(2020)Wolf, Debut, Sanh, Chaumond, Delangue, Moi, Cistac,
  Rault, Louf, Funtowicz, Davison, Shleifer, von Platen, Ma, Jernite, Plu, Xu,
  Scao, Gugger, Drame, Lhoest, and Rush}]{wolf-etal-2020-transformers}
Thomas Wolf, Lysandre Debut, Victor Sanh, Julien Chaumond, Clement Delangue,
  Anthony Moi, Pierric Cistac, Tim Rault, Rémi Louf, Morgan Funtowicz, Joe
  Davison, Sam Shleifer, Patrick von Platen, Clara Ma, Yacine Jernite, Julien
  Plu, Canwen Xu, Teven~Le Scao, Sylvain Gugger, Mariama Drame, Quentin Lhoest,
  and Alexander~M. Rush. 2020.
\newblock \href {https://www.aclweb.org/anthology/2020.emnlp-demos.6}
  {Transformers: State-of-the-art natural language processing}.
\newblock In \emph{Proceedings of the 2020 Conference on Empirical Methods in
  Natural Language Processing: System Demonstrations}, pages 38--45, Online.
  Association for Computational Linguistics.

\bibitem[{Wu et~al.(2019)Wu, Ren, Luo, and
  Sun}]{wu-etal-2019-hierarchical-reinforced}
Chen Wu, Xuancheng Ren, Fuli Luo, and Xu~Sun. 2019.
\newblock \href {https://doi.org/10.18653/v1/P19-1482} {A hierarchical
  reinforced sequence operation method for unsupervised text style transfer}.
\newblock In \emph{Proceedings of the 57th Annual Meeting of the Association
  for Computational Linguistics}, pages 4873--4883, Florence, Italy.
  Association for Computational Linguistics.

\bibitem[{Xu et~al.(2018)Xu, Sun, Zeng, Zhang, Ren, Wang, and
  Li}]{xu-etal-2018-unpaired}
Jingjing Xu, Xu~Sun, Qi~Zeng, Xiaodong Zhang, Xuancheng Ren, Houfeng Wang, and
  Wenjie Li. 2018.
\newblock \href {https://doi.org/10.18653/v1/P18-1090} {Unpaired
  sentiment-to-sentiment translation: A cycled reinforcement learning
  approach}.
\newblock In \emph{Proceedings of the 56th Annual Meeting of the Association
  for Computational Linguistics (Volume 1: Long Papers)}, pages 979--988,
  Melbourne, Australia. Association for Computational Linguistics.

\bibitem[{Yi et~al.(2020)Yi, Liu, Li, and Sun}]{xiaoyuan-ijcai}
Xiaoyuan Yi, Zhenghao Liu, Wenhao Li, and Maosong Sun. 2020.
\newblock Text style transfer via learning style instance supported latent
  space.
\newblock In \emph{Proceedings of the Twenty-Ninth International Joint
  Conference on Artificial Intelligence, {IJCAI-20}}, pages 3801--3807.

\bibitem[{Zhang et~al.(2018)Zhang, Ren, Liu, Wang, Chen, Li, Zhou, and
  Chen}]{Zhirui-2018}
Zhirui Zhang, Shuo Ren, Shujie Liu, Jianyong Wang, Peng Chen, Mu~Li, Ming Zhou,
  and Enhong Chen. 2018.
\newblock \href {http://arxiv.org/abs/1808.07894} {Style transfer as
  unsupervised machine translation}.
\newblock \emph{arXiv preprint, arXiv: 1808.07894}.

\bibitem[{Zhou et~al.(2020)Zhou, Chen, Liu, Xiao, Su, Guo, and
  Wu}]{zhou-etal-2020-exploring}
Chulun Zhou, Liangyu Chen, Jiachen Liu, Xinyan Xiao, Jinsong Su, Sheng Guo, and
  Hua Wu. 2020.
\newblock Exploring contextual word-level style relevance for unsupervised
  style transfer.
\newblock In \emph{Proceedings of the 58th Annual Meeting of the Association
  for Computational Linguistics}, pages 7135--7144.

\end{thebibliography}
\bibliographystyle{acl_natbib}

\clearpage
\appendix
\onecolumn
\section{\Large Appendices: \\~ \\ }
\label{sec:appendix}

\setcounter{table}{0}
\renewcommand{\thetable}{A.\arabic{table}}
\setcounter{figure}{0}
\renewcommand{\thefigure}{A.\arabic{figure}}

\bigskip

This appendices include: 1) detailed results for the different steps of the pipeline (\ref{append:ablation-study}); 2) detailed evaluation results of using four human references on YELP (\ref{append:results-four-ref}); 3) example outputs for the different steps of the pipeline (\ref{append:example-outputs-step-all}); 4) example outputs for existing systems we compare to, and our best models (\ref{append:output-comparison}); 5) sample examples for further pre-training (\ref{append:sample-para}); 6) sample examples of human reference on YELP (\ref{append:sample-ref}) \vspace*{0.5cm}.

\subsection{Detailed results for the different steps of the pipeline}
\label{append:ablation-study}
\begin{table*}[ht]
\centering
\resizebox{\linewidth}{!}{%
\begin{tabular}{l|ccccc|ccccc}
\toprule[2pt]
\multicolumn{1}{r|}{\sc Dataset~~~} & \multicolumn{5}{c|}{\textsc{GYAFC (Formality Transfer)}} & \multicolumn{5}{c}{\textsc{YELP (Polarity Swap)}}\\
\hline
 \textsc{Model} & BLEURT & COMET & BLEU & ACC & HM & BLEURT & COMET & BLEU & ACC & HM \\
 
 \toprule
  Original BART & -0.116 & 0.242 & 0.414 & 0.333 & 0.369 & -0.388 & -0.146 & 0.309 & 0.022 & 0.041\\
  \hline
  \rowcolor{LightGray}
 \multicolumn{11}{c}{STEP~1: Further pre-training}\\ \hline
 Further pre-trained BART using whole dataset & 0.012 & 0.209 & 0.420 & 0.357 & 0.386 & -0.412 & -0.282 & 0.179 & 0.040 & 0.065\\
 Further pre-trained BART using subset & 0.011 & 0.225 & 0.441 & 0.693 & 0.539 & -0.347 & -0.178 & 0.247 & 0.166 & 0.199\\
 Further pre-trained BART using synthetic data & - & - & - & - & - & -0.321 & -0.074 & 0.326 & 0.189 & 0.239\\
 \hline
 \rowcolor{LightGray}
 \multicolumn{11}{c}{STEP~2: IBT + Rewards}\\ \hline
 IBT (original BART) & -0.010 & 0.292 & 0.507 & 0.836 & 0.631 & -0.229 & -0.017 & 0.298 & 0.826 & 0.438\\
 IBT (Further pre-trained BART using whole dataset)  & \textbf{0.048} & 0.319 & 0.550 & 0.907 & 0.685 & -0.192 & -0.041 & 0.252 & 0.854 & 0.389\\
 IBT (Further pre-trained BART using subset) & 0.041 & 0.318 & 0.553 & \textbf{0.932} & \textbf{0.694} & -0.176 & 0.026 & 0.295 & 0.853 & 0.438\\
  IBT (Further pre-trained BART using synthetic data) & - & - & - & - & - & -0.246 & -0.035 & 0.302 & 0.884 & 0.450\\
 \midrule
 IBT + SC0 + SC1 + BLEU & 0.033 & 0.313 & 0.552 & 0.929 & 0.693 & -0.187 & 0.001 & 0.285 & 0.860 & 0.428\\
 IBT + SC0 + SC1 + BLEURT & 0.041 & 0.320 & 0.551 & 0.925 & 0.691 & \textbf{-0.149} & 0.031 & 0.295 & 0.784 & 0.429 \\
 IBT + SC1 + BLEU + BLEURT & 0.024 & \textbf{0.321} & 0.544 & 0.928 & 0.686 & -0.195 & -0.016 & 0.286 & 0.881 & 0.432\\
 IBT + SC0 + BLEU + BLEURT & 0.039 & 0.318 & 0.555 & 0.873 & 0.679 & -0.176 & 0.039 & \textbf{0.331} & 0.500 & 0.398\\
 IBT + SC0 + SC1 & 0.027 & 0.314 & 0.550 & \textbf{0.932} & 0.692 & -0.208 & -0.028 & 0.279 & 0.859 & 0.421\\
 IBT + BLEU + BLEURT & 0.036 & 0.318 & 0.552 & 0.857 & 0.671 & -0.204 & 0.017 & \textbf{0.331} & 0.413 & 0.367\\
 IBT without reward & 0.032 & 0.319 & 0.551 & 0.849 & 0.668 & -0.181 & 0.037 & \textbf{0.331} & 0.489 & 0.395\\
 \hline
 \rowcolor{LightGray}
 \multicolumn{11}{c}{STEP~3: Offline training (Model used: original BART + Rewards)}\\ \hline
 Trained with high-quality pairs & 0.030 & \textbf{0.321} & \textbf{0.560} & 0.904 & 0.692 & -0.183 & \textbf{0.046} & 0.316 & \textbf{0.887} & \textbf{0.466}\\
 Trained with subset of paraphrase data  & 0.012 & 0.229 & 0.455 & 0.783 & 0.576 & -0.338 & -0.221 & 0.215 & 0.457 & 0.292\\
\bottomrule[2pt]
\end{tabular}}
\caption{\label{tab:all-ablation-study}
Detailed results for the different steps of the pipeline. Note that (i) SC0 represents the SC reward is used in the supervised training process using pseudo-parallel data, and SC1 is in the process of generating pseudo-parallel data.
}
\end{table*}

\subsection{Detailed evaluation results of using four human references on YELP}
\label{append:results-four-ref}

\begin{table*}[ht]
\centering
\small
\begin{tabular}{l|ccccc}
\toprule[1.3pt]
 \textsc{Model} & BLEURT & COMET & BLEU & ACC & HM \\
 \toprule
 Input Copy & -0.337 & -0.033 & \textbf{0.640} & 0.019 & 0.037\\
 DualRL \citep{fuli-2019} & -0.281 & -0.080 & 0.550 & 0.894 & 0.681\\
 Style-Transformer \citep{dai-etal-2019-style} & -0.390 & -0.158 & 0.553 & 0.857 & 0.672\\
 DGST \citep{li-etal-2020-dgst} & -0.337 & -0.131 & 0.520 & 0.781 & 0.624\\
 StyIns \citep{xiaoyuan-ijcai} & -0.487 & -0.280 & 0.489 & \textbf{0.924} & 0.640\\
 Zhou's \citep{zhou-etal-2020-exploring} & -0.162 & 0.090 & 0.608 & 0.865 & 0.714\\
 \hline
 Ours & \textbf{-0.053} & \textbf{0.192} & 0.610 & 0.887 & \textbf{0.723}\\
\bottomrule[1.3pt]
\end{tabular}
\caption{\label{tab:results-yelp}
Automatic evaluation results using four human references on YELP. 
}
\end{table*}

\clearpage
\subsection{Example outputs for the different steps of the pipeline}
\label{append:example-outputs-step-all}

\begin{table*}[ht]
\resizebox{\linewidth}{!}{%
\centering
\small
\begin{tabular}{l|l|cccc}
\toprule[2pt]
  \textsc{Model} & \ \makecell[c]{\textsc{Informal} $\rightarrow$ \textsc{Formal}} & BLEU & BLEURT & COMET & Style Confidence\\
  \toprule
  Source & So if you're set on that, that's the way to go!! & \multicolumn{4}{c}{-}\\
  Reference 1 & If you are set on that, that is the way to go. & \multicolumn{4}{c}{-}\\
  Reference 2 & If that is your decision, then that is what you should do. & \multicolumn{4}{c}{-}\\
  Reference 3 & So that is the way to go if you are set on that. & \multicolumn{4}{c}{-}\\
  Reference 4 & If you are set on that, then that is the way to go. & \multicolumn{4}{c}{-}\\
  \hline
  M0   & so if you're set on that, that's the way to go!! & 0.417 & 0.175 & 0.568 & 0.000\\
  M1.1 & so if you want to do this, this is the way to go! & 0.301 & 0.204 & 0.354 & 0.003\\
  M1.2 & If you want to do this, this is the way to go. & 0.416 & 0.339 & 0.433 & 0.855\\
  M2.1 & So if you're set on that, that is the way to go. & 0.763 & 0.525 & 0.689 & 0.179\\
  M2.2 & So, if you are set on that, then that is the way to go. & \textbf{0.884} & 0.456 & 0.722 & \textbf{0.880}\\
  M3.1 & So if you are set on that, that is the way to go. & 0.541 & \textbf{0.941} & \textbf{0.734} & 0.617\\
  M3.2 & If you're on board, that's the way to go. & 0.352 & 0.200 & 0.311 & 0.552\\
  
  \toprule
  Source & (saying sorry to him without commiting a mistake is humilation to ur self...AND DON'T EVER GET WEAK!!) & \multicolumn{4}{c}{-}\\
  Reference 1 & Do not apologize if you have not made a mistake. & \multicolumn{4}{c}{-}\\
  Reference 2 & Don't say sorry to him unless you've actually made a mistake. & \multicolumn{4}{c}{-}\\
  Reference 3 & You will just humiliate yourself if you apologize to him without committing a mistake. Do not get weak! & \multicolumn{4}{c}{-}\\
  Reference 4 & Avoid being weak by not saying sorry for something you did not do. & \multicolumn{4}{c}{-}\\
  \hline
  M0   & (saying sorry to him without commiting a mistake is humilation to ur self...and don't ever get weak!!) & 0.100 & -0.985 & -0.452 & 0.000\\
  M1.1 & (saying sorry to him without apologizing for a mistake is humilation to your self... and don't ever get weak & 0.095 & -0.697 & -0.414 & 0.002\\
  M1.2 & I'm sorry, but to apologize without making a mistake is humilation to my self, and don't ever get weak & 0.031 & \textbf{-0.648} & -0.608 & 0.694\\
  M2.1 & (Saying sorry to him without commiting a mistake is humilation to your self. Please don't ever get weak. & 0.089 & -0.661 & \textbf{-0.247} & \textbf{0.980}\\
  M2.2 & Saying sorry to him without commiting a mistake is humilation to your self and do not ever get weak. & 0.099 & -0.655 & -0.377 & 0.758\\
  M3.1 & Saying sorry to him without commiting a mistake is humilation to your self and do not ever get weak. & 0.099 & -0.655 & -0.377 & 0.758\\
  M3.2 & Saying sorry to him without making a mistake is humilation to your self, and don't ever get weak! & \textbf{0.108} & -0.652 & -0.400 & 0.083\\
  \toprule[1.5pt]
  \textsc{Model} & \ \makecell[c]{\textsc{Positive $\rightarrow$ Negative}} & BLEU & BLEURT & COMET & Style Confidence\\
  \toprule
  Source    & the staff are all super friendly and on top of there jobs.       & \multicolumn{4}{c}{-}\\
  Reference & the staff are not friendly and not on top of their jobs. & \multicolumn{4}{c}{-}\\
  \hline
  M0 &   the staff are all super friendly and on top of there jobs.        & 0.163 & -0.561 & -0.169 & 0.000\\
  M1.1 & all the staff are very friendly and they're doing their jobs well.& 0.107 & -0.571 & -0.301 & 0.003\\
  M1.2 & the staff are all super friendly and on top of each same jobs.    & 0.149 & -0.662 & -0.507 & 0.000\\
  M1.3 & the staff are all super unfriendly and on top of there jobs.      & 0.151 & -0.239 & 0.095  & \textbf{1.000}\\
  M2.1 & the staff are all super rude and on top of there jobs.            & 0.151 & -0.513 & 0.048  & \textbf{1.000}\\
  M2.2 & the staff are all super rude and on top of there jobs.            & 0.151 & -0.513 & 0.048  & \textbf{1.000}\\
  M2.3 & the staff are all super rude and on top of there jobs.            & 0.151 & -0.513 & 0.048  & \textbf{1.000}\\
  M3.1 & the staff are not super friendly or on top of there jobs.         & \textbf{0.320} & \textbf{0.322}  & \textbf{0.621}  & \textbf{1.000}\\
  M3.2 & the staff are so friendly and they're doing their jobs.           & 0.148 & -0.663 & -0.326 & 0.001\\
  \toprule
  Source & very good brunch, was impressed with selection and quality. & \multicolumn{4}{c}{-}\\
  Reference & the brunch was bad, with little selection & \multicolumn{4}{c}{-}\\
  \hline
  M0   & very good brunch, was impressed with selection and quality.                 & 0.028 & -1.134 & -0.400 & 0.000\\
  M1.1 & it was a very good brunch, i was impressed by the selection and the quality.& 0.017 & -0.714 & -0.497 & 0.000\\
  M1.2 & very good brunch, was impressed with the selection and quality.             & 0.027 & -1.101 & -0.441 & 0.000\\
  M1.3 & very bad brunch, was impressed with selection and quality.                  & 0.030 & -1.100 & -0.118 & 0.778\\
  M2.1 & very mediocre brunch, was disappointed with selection and quality.          & 0.028 & \textbf{-0.367} & 0.160  & \textbf{1.000}\\
  M2.2 & very disappointing brunch, was disappointed with selection and service.     & 0.028 & -0.532 & -0.020 & \textbf{1.000}\\
  M2.3 & very mediocre brunch, was disappointed with selection and quality.          & 0.028 & \textbf{-0.367} & -0.160 & \textbf{1.000}\\
  M3.1 & very bad brunch, was disappointed with selection and quality.               & \textbf{0.030} & -0.495 & \textbf{0.236}  & \textbf{1.000}\\
  M3.2 & it was a very good brunch, i was impressed with the food and the service.   & 0.017 & -0.865 & -0.649 & 0.000\\
\bottomrule[2pt]
\end{tabular}}
\caption{\label{tab:example-outputs-step-all}
Example outputs for the different steps of the pipeline.
}
\end{table*}

\subsection{Example outputs for existing systems we compare to, and our best models}
\label{append:output-comparison}

\begin{table*}[h]
\resizebox{\linewidth}{!}{%
\centering
\small
\begin{tabular}{l|l|l}
\toprule
 \textsc{Systems} & \makecell[c]{\textsc{Informal} $\rightarrow$ \textsc{Formal}} & \makecell[c]{\textsc{Negative} $\rightarrow$ \textsc{Positive}}\\
 \hline
 Input & i hardly everrr see him in school either usually I see hima t my brothers basketball games. & so, no treatment and no medication to help me deal with my condition.  \\
 \hline
 DualRL & I \textcolor{red}{recognize him} see him in school either I usually \textcolor{red}{see my brothers}. & so, great treatment and \textcolor{red}{great} help me deal with my condition. \\
 StyIns & I \textcolor{red}{would not} see him in school either \textcolor{red}{because} I see \textcolor{red}{to profess} my brothers basketball games. & so, great \textcolor{red}{service} and great \textcolor{red}{location} to help me deal with my condition. \\
 Zhou's & I hardly \textcolor{red}{everrr} see him in school either. see I \textcolor{red}{’``}my brothers games. & so, great treatment and \textcolor{red}{no} medication to help me deal with my condition. \\
 Ours & I hardly ever see him in school, but usually I see him at my brothers basketball games. & so great treatment and great medication to help me deal with my condition. \\
 \hline
 Human & I hardly ever see him in school, usually I see him when I go to my brother 's basketball games. & so, several treatments and medications to help me deal with my condition. \\
\bottomrule
\end{tabular}}
\caption{\label{tab:sample-output}
Example outputs for existing systems we compare to, and our best models. Improperly generated words/phrases are in red. We can observe that: 1) there are still some informal/negative expressions in the generated sentences of previous systems like \citet{zhou-etal-2020-exploring}'s. 2) Some systems introduce noise in the generated sentences and fail in preserving content like DualRAL \citep{fuli-2019}'s and StyIns \citep{xiaoyuan-ijcai}'s. On the contrary, our proposed approach is better at changing input sentences into the target style while preserving most style-independent parts. Furthermore, generated sentences of our system are more fluent than previous systems.
}
\end{table*}

\clearpage
\subsection{Sample examples for further pre-training}
\label{append:sample-para}
\begin{table*}[ht]
\resizebox{\linewidth}{!}{%
\centering
\small
\begin{tabular}{l|l|l|l}
\toprule[2pt]
 \makecell[c]{\textsc{Resource}} & \makecell[c]{\textsc{Task}} & \makecell[c]{\textsc{Informal/Negative}} & \makecell[c]{\textsc{Formal/*Positive}}\\
 \toprule
 \multirow{10}{*}{Paraphrase} & \multirow{5}{*}{Formality} & Now I... I like the interactive side of this job. & I like the interactivity on our work. \\
 & & Y'all got five minutes to finish your smoke. & All of you have five minutes to finish your long smoke.\\
 & & Yeah, well, there's a bridge right here. & Here's one bridge, ten kilometers from there.\\
 & & If they go......they leave the source of power behind. & If they leave, they'll leave their source of strength.\\
 & & Ain't many of us can face it out there sober all the time. & Not too many of us could face this outside as sober.\\
 \cline{2-4}
 &  \multirow{5}{*}{Polarity} & he makes me feel wired. & i... it gives me a funny feeling. \\
 & & there's a room on the empty floor. & there's plenty of free space on the next floor.\\
 & & it's only part of work, you know -- routine clearance. & great. yeah, it's just part of the job, you know... a routine imprint.\\
 & & it was the only thing i liked to buy here. & that is the one thing i actually enjoy buying at this store.\\
 & & wherever it went, it was followed by an admiring crowd of small lassans. & wherever he moved, he was followed by an astonished mob of little lassans.\\
  \toprule
 \multirow{5}{*}{WordNet} & \multirow{5}{*}{Polarity} & some of the worst pizza i've ever had. & some of the best pizza i've ever had.\\
 & & also the inside is dirty as heck. & also the inside is clean as heck.\\
 & & the guy never really even apologized for the mistake. & the guy ever really even apologized for the mistake.\\
 & & wake up or you are going to lose your business. & wake up or you are going to find your business.\\
 & & absolutely the worst care in all my experience with vets! & absolutely the best care in all my experience with vets!\\
\bottomrule[2pt]
\end{tabular}}
\caption{\label{tab:example-output}
Sample examples for further pre-training. * indicates that the sentences are synthetic.
}
\end{table*}

\subsection{Sample examples of human reference on YELP}
\label{append:sample-ref}
\begin{table*}[ht]
\resizebox{\linewidth}{!}{%
\centering
\small
\begin{tabular}{l|l|l}
\toprule[2pt]
 & \makecell[c]{\textsc{Negative} $\rightarrow$ \textsc{Positive}} & \makecell[c]{\textsc{Positive} $\rightarrow$ \textsc{Negative}}\\
 \toprule
Source & ever since joes has changed hands it's just gotten worse and worse. & it's small yet they make you feel right at home. \\
Reference 1 & ever since joes has changed hands it's gotten better and better. & it's small yet they make you feel like a stranger.\\
Reference 2 & ever since joes has changed hands it`s gotten better and better. & it's small and make you feel as small office cabin\\
Reference 3 & since joe changed hands, it has become a better place. & it's small and not friendly at all\\
Reference 4 & ever since joes has changed hands it is getting better & it's small and they make you feel like a stranger\\
 \toprule
Source & there is definitely not enough room in that part of the venue. & i will be going back and enjoying this great place!\\
Reference 1 & there is so much room in that part of the venue & i won't be going back and suffering at this terrible place!\\
Reference 2 & there is definiteley enough room in that part of the venue. & i will not be going back to this terrible place!\\
Reference 3 & there is enough space in that oart of the venue. & i will never come back to this bad place!\\
Reference 4 & there is many room on that venue & i will not be returning to this place and it was unenjoyable\\
 \toprule
Source & so basically tasted watered down. & the drinks were affordable and a good pour.\\
Reference 1 & it didn't taste watered down at all. & the drinks were expensive and half full.\\
Reference 2 & so it's fine because it is not watered down. & the drinks were expensive and a less pour\\
Reference 3 & so basically not tasted watered down. & the drinks were very expensive and a less pour\\
Reference 4 & so basically did not taste watered down. & the drinks were not affordable and a not good pour.\\
 \toprule
Source & she said she'd be back and disappeared for a few minutes. & my husband got a ruben sandwich, he loved it.\\
Reference 1 & she said she'd be back, and didn't disappear at all. & my husband got a reuben sandwich, he hated it.\\
Reference 2 & she said she'd be back and enjoy herself & my husband got a ruben sandwich, he hate it very much.\\
Reference 3 & she said she'd be back and will not disappeared & my husband got a ruben sandwich, he hated it.\\
Reference 4 & she said she'd be back and have a good time & my husband got a ruben sandwich , he did not love it\\
 \toprule
Source & i can't believe how inconsiderate this pharmacy is. & i signed up for their email and got a coupon.\\
Reference 1 & this pharmacy is really considerate. & i signed up for their email and got spam.\\
Reference 2 & i can not imagine how considerate this pharmacy is. & i signed up for their email and got nothing.\\
Reference 3 & the pharmacy was so considerate of me & i signed up for their email and didnt even get offered a deal or anything.\\
Reference 4 & i can not believe how considerate this pharmacy is & i wrote an email and did not obtube anything\\

\bottomrule[2pt]
\end{tabular}}
\caption{\label{tab:example-yelp}
Sample examples of human reference on YELP. The first human reference is provided by \citet{li-etal-2018-delete}, and the 3 additional references are released by \citet{fuli-2019}.
}
\end{table*}

\end{document}